\documentclass[a4paper,11pt]{article}

\usepackage{amsmath,amsthm,amssymb,a4wide}

\newcommand{\N}{\mathbb{N}}

\newcommand{\R}{\mathbb{R}}

\makeatletter
\def\author#1{\gdef\autrun{\def\and{\unskip, }#1}\gdef\@author{#1}}

\makeatother

\def\email#1{email: \href{mailto:#1}{#1} }
\def\subjclass#1{\par\bigskip\noindent\textbf{Mathematics Subject Classification 2020.} #1}
\def\keywords#1{\par\smallskip\noindent\textbf{Keywords.} #1}

\newenvironment{acknowledgments}{\bigskip\small\noindent\textit{Acknowledgments.}}{\par}

\newtheorem{thm}{Theorem}[section]
\newtheorem{cor}[thm]{Corollary}
\newtheorem{lem}[thm]{Lemma}

\theoremstyle{definition}
\newtheorem{defin}[thm]{Definition}
\newtheorem{exa}[thm]{Example}

\numberwithin{equation}{section}

\usepackage[hyperfootnotes=false,colorlinks=true,allcolors=blue]{hyperref}

\begin{document}

% Give an abbreviation of the title for the running page headers.
%\titlerunning{Variational Regularization}

% Here you can enter the full article title.
\title{\textbf{Variational Regularization in Inverse Problems and Machine Learning}}

% Here you can enter the full names of authors separated by \and.
\author{Martin Burger}

% Please do not enter a date.

% Here you can enter the address and email of each author separated by \and by following the example below.
\date{Department Mathematik and Center for Mathematics of Data, Friedrich-Alexander Universit\"at Erlangen-N\"urnberg, Cauerstr. 11, 91058 Erlangen;  \email{martin.burger@fau.de}}
\maketitle

% Here you can enter an optional dedication.
%\begin{dedication}
%To David Hilbert on the occasion of his 158th birthday
%\end{dedication}

% Here you can enter the abstract, the MSC classes, and some keywords.
\begin{abstract}
This paper discusses basic results and recent developments on variational regularization methods, as developed for inverse problems. In a typical setup we review basic properties needed to obtain a convergent regularization scheme and further discuss the derivation of quantitative estimates respectively needed ingredients such as Bregman distances for convex functionals. 

In addition to the approach developed for inverse problems we will also discuss variational regularization in machine learning and work out some connections to the classical regularization theory. In particular we will discuss a reinterpretation of machine learning problems in the framework of regularization theory and a reinterpretation of variational methods for inverse problems in the framework of risk minimization. Moreover, we establish some previously unknown connections between error estimates in Bregman distances and generalization errors.
\subjclass{Primary 65J20
; Secondary 47A52
}
\keywords{Regularization theory, variational methods, inverse problems, machine learning}
\end{abstract}

\section{Introduction}
 
Regularization methods are an approach of fundamental importance in the solution of ill-posed problems. Their main paradigm is to approximate an ill-posed problem by a parametrized family of well-posed problems, with appropriate convergence properties as the regularization parameter and the so-called noise level tend to zero. The noise level is a measure for the size of deterministic and stochastic errors in the data, which are usually the main cause of concern due to the ill-posedness. 

A detailed theory of regularization has been developed in the typical setting of inverse problems, obviously with more precise results in the case of linear forward models  than for nonlinear ones (cf. \cite{benningburger2018,chenghofmann,EnglHankeNeubauer,Schusterbuch,tikhonov} and references therein). Regularization is however not only relevant in inverse problems, similar methods are now routinely used in machine learning, mainly from a practical point of view, with theoretical results often hidden in the statistical theory of generalization (cf. e.g. \cite{jamesetal,goodfellow,kukacka}). The role and objective of regularization is less clear and less developed in the machine learning domain. In this paper we will thus aim to give a unified overview and present some links between the formulations and questions in inverse problems and those in machine learning. We will concentrate on the prominent class of variational regularization methods, which we interpret in a rather broad way.

\section{Regularization Theory}

In order to present the basic ideas of regularization methods in a rather unified way for inverse and machine learning problems, we will first adopt a high-level point of view. Regularization theory is based on the following ingredients:
\begin{itemize}

\item an {\em ideal problem} respectively an {\em ideal solution} $u^*$. We can assume that the ideal problem is given by a map
$\Phi: {\cal V}_D  \rightarrow {\cal U}$, where ${\cal V}_D$ is a space of ideal data and ${\cal U}$ is the space of admissible solutions. The typical analysis is confined to Banach or at least metric spaces.  

\item a space ${\cal V} \supset {\cal V}_D$ of {\em possible data} and a measure of noise between the ideal data $v^* = \Phi^{-1}(u^*) \in {\cal V}_D$ and noisy data $v \in {\cal V}$. In the case of an ill-posed problem, the operator $\Phi$ is not continuous when considered from (a subset of) ${\cal V}$ to ${\cal U}$, it may be continuous on bounded subsets of ${\cal V}_D$ however. The latter leads to the concept of conditional stability (cf. \cite{weidling,wernerhofmann}) and corresponding stability estimates.

\item a family of continuous, possibly multivalued, maps $\Phi_\alpha: {\cal V} \rightarrow {\cal P}({\cal U})$, $\alpha \in {\cal A}$, such that for a sequence $(v_n) \subset {\cal V}$ converging to $v^* \in {\cal V}_D$, there exists a parameter sequence $\alpha_n$ such that there is $u_n \in \Phi_{\alpha_n}(v_n)$ converging to $u^*$ (in a suitable metrizable topology, possibly weak or weak-star on bounded sets in the Banach space case). Sometimes the notion of convergence is restricted to subsequences.

\end{itemize} 

To make these notions more concise we will discuss them in the setting of inverse problems as well as machine learning subsequently.

\subsection{Inverse Problems} 

In the typical case of inverse problems, there is first a (continuous) forward operator $F: {\cal U} \rightarrow {\cal V}$, which is typically not invertible and if it is on a subset of ${\cal V}$, the inverse is discontinuous. The set of ideal data is a subset of $F({\cal U})$, and there the multivalued operator
$$ \Phi_0: {\cal V}_D \rightarrow {\cal P}({\cal U}),  v  \mapsto F^{-1}(v) $$
can be defined. In order to obtain a unique (generalized) inverse, a further selection operator 
$ \Sigma:  {\cal P}({\cal U}) \rightarrow {\cal U}$ is defined to obtain $\Phi:= \Sigma \circ \Phi_0$. Let us mention that there are standard examples of the selection operator such as the minimum norm solution, but often this issue is treated in a hidden or unprecise way. We refer to \cite{benningburger2018} for a detailed discussion of selection operators in inverse problems. 

The standard notion of noise is the perturbation of the data, i.e. $v-v^*$, either as a deterministic or a stochastic quantity. The norm of $v-v^*$ in the Banach space ${\cal V}$ (or the expectation of some power of the norm) serves as a definition of the noise level.

The solution of the inverse problem can then be cast as the solution of the ill-posed operator equation
$$ F(u) = v$$
or, as the minimization of 
\begin{equation}
D(u) =   L(F(u),v),
\end{equation} 
where $L$ is some distance measure between the predicted data $F(u)$ and the measured data $v$. If statistical information about the noise is available or the forward model contains other stochastic elements, ${ L}$ is typically a negative log-Likelihood functional. 

As mentioned above,  regularization methods are families of multivalued operators 
$ \Phi_\alpha: {\cal V} \rightarrow {\cal P}({\cal U}) $, 
in most cases the parameter domain ${\cal A}$ is a subset of the positive real numbers. 
The well-posedness of $\Phi_\alpha$ is characterized by some set-valued continuity, e.g. if $u_n \rightarrow u$ then 
$\Phi_\alpha(u_n)$ contains a convergent subsequence and each limit $v$ of a convergent subsequence satisfies $v \in \Phi_\alpha(u)$. In most cases the regularization operator satisfies a stronger stability estimate of the form 
\begin{equation} 
\label{eq:phialphastability} d_U (u_1,u_2) \leq C_\alpha d_V(v_1,v_2)  \qquad \forall u_1 \in \Phi_\alpha(v_1), u_2 \in \Phi_\alpha(v_2),\end{equation}
where $d_U$ and $d_V$ are appropriate distance measures (that may be degenerate in the sense that $d_U(u_1,u_2)$ can vanish also if $u_1 \neq u_2$).

Regularization methods are constructed along several different paradigms:
\begin{itemize}
\item Data smoothing or mollifier methods, which are of the form $\Phi_\alpha = F^{-1} \circ M_\alpha$, where $M_\alpha: {\cal V} \rightarrow {\cal V}_D$ is family of mollifying (smoothing) operator that into an appropriate subspace of ${\cal V}$ on which there exists a continuous inverse of $F$. In order to obtain suitable regularization methods a quite detailed characterization of the forward operator is needed in order to be sure to construct a mollification to the right subspace. Consequently such methods became popular for inverse problems with well-understood  forward operators such as tomography (cf. \cite{louis1996approximate,louismaass}).

\item Direct approximation of the operator $F$ by continuously invertible operators (cf. \cite{EnglHankeNeubauer,kirsch,lavrentiev,tikhonov} and references therein). The construction of approximations is usually done only in the case of linear forward operators based on modifying (small) singular vectors or by approximating the normal equation, i.e. $F^* F$. The latter is however related to the minimization of the least-squares function $\Vert F(u) - v \Vert^2$ and can thus be viewed as a variational method. Another approach modifying the forward operator is discretization, the regularization parameter thus being related to the discretization fineness.

\item Variational methods are based on a perturbation of the likelihood minimization, $\Phi_\alpha$ mapping $v$ to the set of minimizers of
$$  D_\alpha(u) =  { L}(F(u),v) + \alpha J(u) $$
for some regularization functional $J$ that introduces the needed compactness properties for the existence of minimizers and $\alpha \in \R_+$being the regularization parameter (cf. \cite{benningburger2018,Schusterbuch}).

\item Iterative regularization methods use a well defined iteration method such as a fixed-point iteration or some descent scheme for the likelihood minimization to define an approximation of the inverse of $F$, with the iteration number $\alpha \in \N$ being the regularization parameter (cf. \cite{burgerkaltenbacherneubauer,EnglHankeNeubauer,kaltenbacherneubauerscherzer,kaltenbacher2009iterative,osher2005iterative}). Since the majority of iterative methods, in particular in the nonlinear case, are iterative methods for variational problems, there is an intimate connection to variational regularization methods.

\item Learned regularization methods are of increasing relevance recently (cf. \cite{arridgeetal,benningburger2018} and references therein), which are categorized into supervised and semi-supervised approach. The supervised approach tries to learn the regularization operator $\Phi_\alpha$ directly from a collection of pairs of training data
$(u_i,v_i)$, e.g. by approximation with a deep neural network. Consistent data pairs are however difficult to obtain in many inverse problems, in particular with realistic input data $u_i$ and realistic noise in $v_i$. The alternative semi-supervised approach mainly works on suitable solutions $u_i$, e.g. images for reconstruction tasks, and tries to learn a more conventional regularization approach, e.g. the regularization functional $J$ in a variational regularization methods. With certain restrictions such as convex networks those become accessible for theoretical arguments of regularization theory.
\end{itemize}

Besides providing a well-posed problem for fixed $\alpha$, which often requires some advanced analysis itself (e.g. existence of minimizers for variational problems), a major goal of regularization theory is to study the convergence of regularized solutions. While a qualitative convergence theory can be developed under generic conditions, it is well-known that a quantitative theory will rely on additional assumptions
on the ideal solution $u^*$ due to the underlying ill-posedness. To understand the possibility to derive such estimates and the used assumptions from a generic point of view, let us consider a sequence of data $v_n \rightarrow v^*$ and a parameter choice $\alpha_n$, assuming that $\alpha_n$ is a nonnegative scalar sequence converging to zero (e.g. the regularization parameters in a variational regularization method or $\alpha_n = \frac{1}{k_n}$ with $k_n$ the maximal iteration number in an iterative regularization method). Now assume that the stability estimate \eqref{eq:phialphastability} holds and that $u^*$ satisfies a {\em range condition} for the regularization operator (cf. \cite{benningburger2018}).

\begin{defin}
An element $u^* \in {\cal U}$ is said to satisfy a \emph{range condition for the regularization operator} $\Phi_\alpha$ if
 for all $\alpha$ there exists $v_\alpha^*$ such that $u^* \in \Phi_\alpha(v_\alpha^*).$
\end{defin} 

Under a range condition we can write 
$$ u_n - u^* \in \Phi_{\alpha_n}(v_n) - \Phi_{\alpha_n}(v_{\alpha_n}^*) $$
and exploit the stability estimate \eqref{eq:phialphastability} to obtain
$$  d_U (u^*,u_n) \leq C_{\alpha_n} d_V(v_{\alpha_n}^*,v^*) . $$
Thus, if we can control the range condition in the sense that we can construct an element $v_{\alpha_n}^*$ out of $v^*$ such that the distance can be estimated, we directly obtain an error estimate. This will be made more precise in the next section on variational regularization methods. 

\subsection{Learning and Risk Minimization} 

In the typical case of machine learning problems (cf. \cite{jamesetal,mohri}) we are given (randomly sampled) input samples $x_i \in {\cal X}$ and output samples $y_i \in {\cal Y}$, $i=1,\ldots,N$  and want to infer a parametrized map $f_\theta : {\cal X} \rightarrow {\cal Y}$ reasonably reproducing these training data and generalizing further to other data of the same kind. These properties are frequently obtained from risk minimization arguments. Given a loss $\ell$ measuring deviations in the output space, the empirical risk is given by
$$ \hat R(\theta) = \frac{1}N \sum_{i=1}^N \ell(f_\theta(x_i),y_i) $$
and approximate solutions are constructed as approximate minimizers of $\hat R$, e.g. via variational regularization methods minimizing
$$ D_\alpha(\theta) = \hat R(\theta)  + \alpha J(\theta) $$
or by iterative methods such as the gradient descent 
$$ \theta^{k+1}  = \theta^k - \tau^k \hat R'(\theta^k) $$
or even more often by stochastic gradient descent, where the term implicit regularization is common (cf. \cite{neyshabur}). 

Generalization is usually measured by the behaviour on the population risk, i.e. 
$$ R(\theta) = {\mathbb E}_{(x,y) \sim {\mathbb P}} (\ell(f_\theta(x),y)) ,$$
in particular the generalization error defined by
$$ G(\theta) = R(\theta) - \hat R(\theta) ,$$
evaluated at a regularized solution.
Note that the generalization error is actually a random variable depending on the samples $(x_i,y_i)_{i=1}^N$, hence 
it is relevant to consider its distribution among the random sampling.

The ideal model could be defined in two ways, depending on what variable is identified to be the relevant one. In any case the ideal solution is perceived as a minimizer of the population risk, however one could define $u^*$ as the optimal parameter value or the optimal function. Thus we are led to the following cases
\begin{itemize}

\item[(i)] The first case, corresponding to classical approaches in statistics such as regression, is to define ${\cal U}$ as the set of possible parameters, genuinely a finite-dimensional space (with few generalizations to infinite-dimensional models recently, cf.\cite{stuart,nelsen}). Thus, the ideal solution is given by
$$ \theta^* \in \text{arg}\min_{\theta\in {\cal U}} R(\theta) , $$

\item[(ii)] The second case rather corresponds to the perspective of modern learning theory, it extends the population risk to some function class ${\cal F}$, and computes for $f \in {\cal F}$
$$ S(f) = {\mathbb E}_{(x,y) \sim {\mathbb P}} (\ell(f(x),y)) .$$
The ideal solution is given by
$$ f^* \in \text{arg}\min_{f \in {\cal F}} S(f). $$

\end{itemize}

Another obvious question in this case is how to define the ideal and perturbed data. We follow a distributional viewpoint and define the ideal data $v^*$ as the data distribution ${\mathbb P}$. Correspondingly, the perturbed data are given by the empirical distribution
$$ {\mathbb P}^N =  \frac{1}N \sum_{i=1}^N \delta_{(x_i,y_i)}, $$
where $\delta_z$ denotes the concentrated measure at $z$. Thus, the noise level becomes a distance between (probability) distributions, standard distances such as the total variation distance or Wasserstein metrics. 

The regularization operator $\Phi_\alpha$ maps from a space of probability distributions to (set of) regularized solutions. Take the variational regularization of minimizing $D_\alpha$ as an example. Then in case (i), $\Phi_\alpha$ is given by
$$ \Phi_\alpha:  {\mathbb P}^N \mapsto \text{arg}\min_\theta D_\alpha (\theta) , $$
while in the second case (ii) we have
$$ \Phi_\alpha:  {\mathbb P}^N \mapsto  \{ f_\theta ~|~ \theta \in \text{arg}\min_\theta D_\alpha (\theta) \}. $$

We finally mention that these models can obviously be generalized, in particular to the case of further data errors in the samples $(x_i,y_i)$.  Then the samples can be considered to be drawn from a distribution ${\mathbb P}'$ and the effective error is not just determined by sampling but also by the distance of ${\mathbb P}$ and ${\mathbb P}'$. 

Thus, we see that regularized learning problems can be reformulated in the language of regularization theory for inverse problems (see also \cite{burgerengl,rosasco}). In turn we will see that many inverse problems can be reformulated as risk minimization problems, in particular if there is additional sampling of measurement points.

\subsection{Risk Minimization Formulation of Inverse Problems}

Many inverse problems are dealing with data being functions of a variable $x$, e.g. in integral equations of the first kind or tomography, where $x$ is a set of distances and angles (cf. \cite{natterer}). Denoting the unknown of the inverse problem by $\theta$ we thus obtain $F(\theta)$ as function of $x$ and denote $f(x;\theta)  = F(\theta)(x)$. Moreover, standard log-likelihood functionals in this setting are of the form
$$ L(F(\theta),v) = \int_\Omega \ell(F(\theta)(x),v(x)) ~dx $$ 
for some function $\ell$. Thus, choosing ${\cal P} = {\cal L}_\Omega \delta_{v(x)} $, where ${\cal L}_\Omega$ denotes the Lebesgue measure on $\Omega$, we obtain 
$$ L(F(\theta),v) =  {\mathbb E}_{(x,y) \sim {\mathbb P})}(\ell(F(\theta)(x),y) = {\mathbb E}_{(x,y) \sim {\mathbb P}}(\ell(f(x;\theta),y).$$
The ideal problem is thus the minimization of the loss for appropriate data $v^*$.

In a practical setting we have a finite sampling of data with additional noise, which we consider to be additive for simplicity in the following. This means the practical data are a finite number $N$ of samples $y_i = F(\theta)(x_i) + n_i$, where $n_i$ are the noise samples drawn from some distribution. The practical distribution of samples and data is of the form
$$ {\mathbb P}^N = \frac{1}{N} \sum_{i=1}^N  \delta_{x_i} \otimes  \delta_{F(\theta^*)(x_i)+n_i}  $$
where the $x_i$ are drawn from a prior distribution (usually a deterministic one) and the $n_i$ are drawn from  the noise distribution
%The continuum counterpart are data of the form $y = A(\theta^*)(x) + n$, with $x$   from a given distribution and $y$ from a conditional distribution upon $x$, which is the push-forward of the noise distribution 

\begin{exa} \label{radonexample}
As a simple example consider the inversion of the Radon transform on a domain $\Omega \subset \R^2$. Then in the standard parametrization  we can choose
$x \in [0,\pi) \times [0,L]$ as the angle and distance to origin of the lines to be integrated on. Correspondingly $F(\theta)(x)$ is the line integral of the density function $\theta$ on the line parametrized by $x$.  Now let $x$ be drawn from the uniform distribution on $[0,\pi) \times [0,L]$, each $n$ be drawn from a Gaussian distribution $G_\sigma$ with zero mean and finite variance. Then the population risk becomes \begin{align*} R(\theta) &= \frac{1}{2 L \pi} \int_{[0,\pi) \times [0,L]} \int_\R | F(\theta)(x) - F(\theta^*)(x) - n|^2~dG_\sigma(n)~dx \\
&=  \frac{1}{2 L \pi} \int_{[0,\pi) \times [0,L]} | F(\theta)(x) - F(\theta^*)(x) |^2~dx 
+ \int_\R n^2~dG_\sigma(n). \end{align*}
Hence, after affine transform with terms independent of $\theta$, the population risk equals the squared $L^2$-distance of the 
Radon transforms of $\theta$ and $\theta^*$, which is the usual data discrepancy ${L}$. The empirical risk on the other hand is of the form
$$ \hat R(\theta) = \frac{1}{2N} \sum_{i=1}^N \vert F(\theta)(x_i) - y_i \vert^2, $$
which is the standard functional minimized in practice. 
\end{exa} 

For a more general noise model one may construct the conditional distribution for $y$ based on using the appropriate push-forward of the noise distribution based on applying the noise to $F(\theta^*)(x)$ and an appropriately chosen loss function. Moreover, errors in the forward model could be included in the stochastic model, which will imply that even in the ideal model the conditional distribution of $y$ given $x$ is not concentrated.
%
%\subsection{Ill-Posedness and Generalization}
 %
%Since we can write well-known ill-posed problems in the same risk minimization form as supervised learning, it seems obvious that may expect the ill-posedness of learning problems. 
%
%
%$$ f(x;\theta) = \frac{1}N \sum_{i=1}^N \theta_i^2 \sigma(\theta_i^1 \cdot x + \theta_i^0). $$
%
%$$ F(x) =  \int \theta^2(a,b) \sigma(a \cdot x + b) ~d\mu(a,b)$$ 
%
%$$ F^* = \text{arg}\min_{F \in {\cal F}} ~ \expec_{(x,y) \sim P}(\ell(F(x),y) )$$ 
%
%$$ \expec_{ P}(\partial_F \ell(F^*(x),y)(F(x) - F^*(x) ) = 0$$ 
%
%
%$$ d(f,F^*) =  \expec_{ P}(\ell(f(x;\theta),y) - \ell(F^*(x),y) - \partial_F \ell(F^*(x),y)(f(x;\theta) - F^*(x) ) )$$
%
%
%\begin{align*}
%d(f,F^*) =&  \expec_{ P}(\ell(f(x;\theta),y)) -  \expec_{ P'}(\ell(f(x;\theta),y))  + \\
         %&  \expec_{ P'}(\ell(f(x;\theta),y)-\ell(F^*(x),y)) + \\
				%& \expec_{ P'}(\ell(F^*(x),y)) - \expec_{ P}(\ell(F^*(x),y))
%\end{align*}
%
%

\section{Variational Regularization}

In the following we present some key steps in the analysis of iterative regularization methods, for the sake of a simpler presentation restricting ourselves to a linear forward model and a quadratic data fidelity in a Hilbert space, i.e. 
\begin{equation} \label{eq:Dalpha}
D_\alpha(u)  = \frac{1}2 \Vert F u - v \Vert^2 + \alpha J(u),
\end{equation} 
where $J: {\cal U} \rightarrow \R \cup \{+\infty\}$ is assumed to be convex and proper. Moreover, we assume ${\cal V}$ to be a Hilbert space and ${\cal U}$ a Banach space being the dual of some Banach space ${\cal W}$, with the additional property that the weak-star topology on ${\cal U}$ is metrizable on bounded sets. The operator $F: {\cal U} \rightarrow {\cal V}$ is assumed to be bounded and the adjoint of a bounded linear operator $E: {\cal V} \rightarrow {\cal W}$. With abuse of notation we shall write $F^* = E.$ Finally, we need some additional property of the regularization functional, we assume that it is the convex conjugate of some other functional $H: {\cal W}\rightarrow \R$, i.e. 
$$ J(u) = \sup_{w \in {\cal W}} \langle u, w \rangle - H(w). $$
Let us mention that convex conjugates are weak-start lower semicontinuous, which is obviously an important property of the functional and can be infered by similar arguments as the weak lower semicontinuity results in \cite{EkelandTemam}.
Finally, a coercivity property is needed to apply weak-star compactness arguments (based on the Banach-Alaoglu theorem), we assume that the sublevel sets 
$$ M_C = \{ u \in {\cal U}~|~J(u) \leq C  \}$$
are bounded in ${\cal U}$ for $C > 0$. The final property we need is that $J$ is bounded below, we can assume directly that $J$ is nonnegative.

There are various important examples in literature motivating the above model and assumptions. A popular and reasonably easy to compute approach is classical Tihonov-Phillips regularization with ${\cal U}$ being a Hilbert space and
$$ J(u) = \frac{1}2 \Vert u \Vert^2. $$
Possibly the most prominent example with a variety of applications is total variation regularization (cf. \cite{chambolle2010introduction,tvzoo}), i.e. ${\cal U}=BV(\Omega)$ and 
$$ J(u) = \sup_{g \in C_0^\infty(\Omega)^d, \Vert g \Vert_\infty \leq 1} \int_\Omega u \nabla \cdot g ~d{\cal L}_\Omega, $$
where $\Omega \subset \R^d$ is the domain on which the function to be reconstructed is defined. There are various variants of total variation, including higher order versions, which received considerable attention. Another class of important regularization methods are sparsity-enforcing priors (cf. \cite{RamlauTeschke}), in the simplest setup ${\cal U} = \ell^1$ and
$$ J(u) = \sum |u_i|. $$
An interesting case in deconvolution problems as well as mean-field approaches to learning with neural networks is the continuum variant, the total variation norm of Radon measures (cf. \cite{bredies2013inverse,denoyelle2017support}). Here we have ${\cal U} = {\cal M}(\Omega)$ and 
$$ J(u) = \sup_{w \in C_0(\Omega) } \int_\Omega w ~du .$$

\subsection{Basic Properties of Variational Regularization Methods}

A key result, often found for special cases in literature (cf. e.g. \cite{seidman1989well,tvzoo}) is the existence of a minimizer and some stability, which verifies the well-posedness of the regularization operator 
$\Phi_\alpha(v) := $arg$\min_u D_\alpha(u)$.
\begin{thm}
Under the above assumptions on ${\cal U}$, ${\cal V}$, $F$, and $J$ there exists a minimizer of $D_\alpha(u)$ for every $v \in {\cal V}$ and every $\alpha > 0$. Moreover, if $\alpha > 0$, $v_n \rightarrow v$ and $u_n \in \Phi_\alpha(v_n)$, then there exists a weak-star convergent subsequence $v_{n_k}$ and the limit $u$ of every weak-star convergent subsequence satisfies $u \in \Phi_\alpha(v)$.
\end{thm}

In general no uniqueness can be shown under the above conditions, which is anyway not to be expected for the rather degenerate examples above. 
However, a weaker type of uniqueness can be inferred from the convexity and optimality condition 
$$ F^* (Fu  - v) + \alpha p = 0, \qquad p \in \partial J(u), $$
where $\partial J(u)$ denotes the subdifferential 
$$ \partial J(u) = \{ w \in {\cal U}^* ~|~ J(u) + \langle w, \tilde u -u \rangle \leq J(\tilde u) \quad \forall \tilde u \in {\cal U} \}. $$
From the assumptions on $F$ we see that $F^*$ effectively maps to the predual space ${\cal W}$, thus the subgradients in the optimality condition effectively satisfy $p \in {\cal W}$, which is a weak regularity condition. A key concept needed in the following is the Bregman distance or generalized Bregmandistance (cf. \cite{burgerbregman,kiwiel1997proximal}):

\begin{defin}
Let $J: {\cal U} \rightarrow \R \cup \{+\infty\}$ be a convex proper functional, and $u, \tilde u \in {\cal U}$ with $p \in \partial J(u)$. Then the (generalized) Bregman distance $d_J^p(\tilde u, u)$ is defined by 
$$ d_J^p(\tilde u, u) = J(\tilde u) - J(u) - \langle p, \tilde u -  u \rangle. $$
If $\tilde p \in \partial J(\tilde u)$ the symmetric Bregman distance $d_J^{\tilde p,p}(\tilde u, u)$ is defined by
$$d_J^{\tilde p,p}(\tilde u, u) = \langle \tilde p - p, \tilde u -  u \rangle. $$
\end{defin} 

Now assume that there are two minimizers $u_1$ and $u_2$ of the variational regularization problem, then the difference in optimality conditions yields
$$ F^* F(u_1-u_2) + \alpha (p_1-p_2) = 0 $$
and from a duality product with $u_1-u_2$ we infer
$$ \Vert F (u_1-u_2) \Vert^2 + \alpha d_J^{p_1,p_2}(u_1, u_2) = 0. $$
Hence, by the nonnegativity of both terms we obtain uniqueness of the output value, i.e. $Fu_1 = Fu_2$ as well as a vanishing symmetric
Bregman distance between $u_1$ and $u_2$.

Finally we can turn our attention to convergence properties of the regularization method. For this sake we use an exposition based on $\Gamma$-convergence (cf. \cite{braides}):
\begin{lem}
Let $v_n \rightarrow v^* =  F u^*$ in ${\cal V}$ and $\alpha_n \rightarrow 0$. Then the sequence of functionals $D_{\alpha_n}$ defined by
$$ D_{\alpha_n}(u) = \frac{1}2 \Vert F u - v_n \Vert^2 + \alpha_n J(u) $$
$\Gamma$-converges to
$$ D_0(u) = \frac{1}2 \Vert F u - v^* \Vert^2 $$
with respect to the weak-star topology in ${\cal U}$.
\end{lem}

This kind of convergence is not strong enough to infer convergence of minimizers, in particular since there is no equicoercivity property. To achieve this, we need to rescale the functional, i.e. use $\Gamma$-convergence by development to the next order:
\begin{lem}
Let $v_n \rightarrow v^* =  F u^*$ in ${\cal V}$ and $\alpha_n \rightarrow 0$ such that
$$ \frac{\Vert v_n- v^* \Vert^2}{\alpha_n} \rightarrow 0. $$
Then the sequence of functionals $E_{\alpha_n}$ defined by
$$ E_{\alpha_n}(u) = \frac{1}{2\alpha_n} \Vert F u - v_n \Vert^2 +  J(u) $$
$\Gamma$-converges to
$$ E_0(u) = \left\{ \begin{array}{ll} J(u) & \text{if } Fu = v^* \\ + \infty & \text{else} \end{array} \right. $$
with respect to the weak-star topology in ${\cal U}$.
\end{lem}
 
Let us mention that we obtain divergence, i.e. $E_{\alpha_n}$ converges to the functional identically equal to $+\infty$, if the condition on the parameter choice is violated, i.e. $\lim\inf \frac{\Vert v_n- v^* \Vert^2}{\alpha_n}  > 0$.
Since $E_\alpha \geq J$ and $J$ is coercive, we immediately conclude the equi-coercivity of the sequence $E_{\alpha_n}$.

\begin{cor} \label{cor1}
Let $v_n \rightarrow v^* =  F u^*$ in ${\cal V}$ and $\alpha_n \rightarrow 0$ such that
$$ \frac{\Vert v_n- v^* \Vert^2}{\alpha_n} \rightarrow 0. $$
Moreover, let $u_n$ be a sequence of minimizers of  $D_{\alpha_n}$ (or equivalently $E_{\alpha_n}$), then there exists a subsequence converging
with respect to the weak-star topology in ${\cal U}$ and the limit $u^{**}$ of each weakly convergent subsequence is a minimizer of $E_0$.
Moreover, $J(u_n) \rightarrow J(u^{**})$.
\end{cor}

Corollary \ref{cor1} confirms that indeed the regularization operator defined by $$\Phi_\alpha(v) = \text{arg}\min_u D_\alpha(u)$$ yields a convergent regularization. Let us mention some further direct consequences:
\begin{itemize}

\item If the $J$-minimizing solution is unique, i.e. $u^{**}$ is the unique minimizer of $E_0$, then the whole sequence $u_n$ converges weakly to $u^{**}$. Moreover, if there is $p^{**} \in \partial J(u) \cap {\cal W}$, then due to the convergence of $J$ and the weak star convergence we conclude
$$d_J^{p^{**}}(u_n,u^{**})  \rightarrow 0 . $$

\item If $u^*$ satisfies $Fu^* = v^*$, but is not $J$-minimizing solution  (a minimizer of $E_0$), it cannot be reconstructed by the regularization method, i.e. 
it is not the limit of minimizers of the variational regularization for positive $\alpha$. This is related to the question whether the regularization functional introduces the right type of prior knowledge. If we are interested in  reconstructing a solution like $u^*$ that is not $J$-minimizing, then $J$ is not a suitable choice. 

\item If $J $ is the norm in ${\cal U}$ as in many frequent examples and ${\cal U}$ satisfies a Radon-Riesz property, the previous result indeed implies strong convergence of subsequences. 
\end{itemize} 

The above analysis was based on a deterministic approach, but in a similar way a stochastic theory can be developed, e.g. for a sequence of random variables $v_n$ with variance ${\mathbb E}(\Vert v_n - v^* \Vert^2)$ converging to zero. 

\subsection{Quantitative Estimates}

As mentioned above it is important to derive quantitative estimates between solutions of the regularized problem and ideal solutions, which we   present here based on  using range conditions as sketched above. In the following we denote by $u_\alpha$ a regularized solution, i.e. a minimizer of $D_\alpha$. Due to convexity $u_\alpha \in \Phi_\alpha(v)$ is characterized as the solution of of the optimality condition
$$ F^*(F u_\alpha - v) + \alpha p_\alpha = 0, \qquad p_\alpha \in \partial J(u_\alpha). $$
Taking two such solutions one can establish a stability estimate for the Bregman distance (cf. \cite{benningburger2018}):

\begin{thm}
Let $u_\alpha \in \Phi_\alpha(v)$ and  $\tilde u_\alpha \in \Phi_\alpha(\tilde v)$. Then the estimate
$$ \frac{1}2\Vert Fu_\alpha - F \tilde u_\alpha \Vert^2 + \alpha d_J^{p_\alpha,\tilde p_\alpha}(u_\alpha,\tilde u_\alpha) \leq 
\frac{1}2 \Vert v- \tilde v \Vert^2 $$
holds, where $p_\alpha$ respectively $\tilde p_\alpha$ are the subgradients appearing in the optimality condition for $u_\alpha$ respectively $\tilde u_\alpha$.
\end{thm} 

Now we turn to the range condition, effectively reformulating a result from \cite{burgerosher}:
\begin{lem}
An element $u^* \in {\cal U}$ with $v^* = Fu^*$ satisfies the range condition for the variational regularization operator $\Phi_\alpha$ 
if and only if it satisfies the {\em source condition} 
$$ \exists z^* \in {\cal V}: F^* z^* \in \partial J(u^*). $$
\end{lem} 

The key part of the proof is the explicit construction $v_\alpha^* = v^* + \alpha z^*$, which allows to obtain an estimate of the right-hand side in the error estimate, due to 
$$ \Vert v- v^*_\alpha \Vert \leq \Vert v- v^*  \Vert  + \Vert v^*- v^*_\alpha \Vert = \Vert v- v^*  \Vert + \alpha \Vert z^* \Vert. $$
This leads to the error estimates as derived in \cite{burgerosher}:

\begin{cor} 
Let $u_\alpha \in \Phi_\alpha(v)$ and let $v^* = Fu^*$, with $u^*$ satisfying the source condition $p^*=F^* z^* \in \partial J(u^*)$. Then the estimate 
$$ \frac{1}2\Vert Fu_\alpha - F u^* \Vert^2 + \alpha d_J^{p_\alpha,p^*}(u_\alpha,u^*) \leq \Vert v- v^*  \Vert^2 + {\alpha^2} \Vert z^* \Vert^2. $$
\end{cor}

In the error estimate we see again the condition on the choice of $\alpha$ needed for the convergence of regularization methods. While the estimate on the output error $\Vert Fu_\alpha - F u^* \Vert$ is uniform in $\alpha$, the effective estimate for the Bregman distance is of the form 
$$  d_J^{p_\alpha,p^*}(u_\alpha,u^*) \leq \frac{\Vert v- v^*  \Vert^2}\alpha + {\alpha} \Vert z^* \Vert^2, $$
which is small again only if $\alpha$ and the quotient $ \frac{\Vert v- v^*  \Vert^2}\alpha$ are small.  

One also observes a bias-variance decomposition inherent in the estimate, even more clear when we assume an underlying stochastic noise model, i.e., $v$ is a random variable. Without systematic errors in the measurements, we have ${\mathbb E}(v) = v^*$ and hence 
$$  {\mathbb E} ( d_J^{p_\alpha,p^*}(u_\alpha,u^*)) \leq \frac{{\mathbb E}(\Vert v- v^*  \Vert^2)}\alpha + {\alpha} \Vert z^* \Vert^2. $$
The measure on the left-hand side is the natural generalization of the mean-squared error to the case of convex variational regularization, the right-hand side is composed of the data variance and the bias term $\Vert z^* \Vert^2$, scaled by the regularization parameter.

Let us mention that the above estimates in Bregman distances lead to estimates in norms if $J$ satisfies strong convexity conditions (cf. \cite{Schusterbuch}).  In the case of not strictly convex functionals the Bregman distance can vanish even if $u_\alpha \neq u^*$, e.g. in total variation regularization if they differ by a change of contrast $u_\alpha = h(u)$ with a monotone function $h$, but rather measures a deviation of the discontinuity sets (cf. \cite{benningburger2018,tvzoo}). In such cases the multivaluedness of the subdifferential can even be an advantage that needs to be exploited, since we do not have just a single estimate, but actually an estimate for each $p^*$  satisfying a source condition. Estimates for other quantities can then derived from the Bregman distance estimates by optimizing over the possible $p^*$ and the associated source elements $z^*$ (respectively their norm appearing in the error estimates. An example are estimates for total variation regularization for piecewise constant functions, it has been shown already in  \cite{burgerosher} how the total variation of $u_\alpha$ away from the discontinuity set of $u^*$ can be estimated by choosing appropriate subgradients.

Again the above type of conditions and estimates are the canonical ones, but can be developed much farther 
(cf. e.g. \cite{benningburger2011,flemming2010new,flemming2013variational,grasmair2011linear,hofmannetal2007,hofmannmathe,hohageweidling,Resmerita2005,resmerita06,weidling}). The first issue is the question of having better estimates under stronger conditions, and a typical example is an improved source condition $p^* =  F^* F \eta^* \in \partial J(u^*)$ for some $\eta^* \in {\cal U}$. In this case the element $\eta^*$ can be used to construct an approximate solution $u_\alpha^* = u^* - \alpha \eta^*$ instead of approximate data for a range condition. This was carried out in \cite{Resmerita2005} (see also \cite{grasmaierhigher}) to obtain the estimate
$$ d_J^{ p^*}(u_\alpha,u^*) \leq d_J^{ p^*}(u^* - \alpha \eta^*, u^*) + \frac{\Vert v- v^*  \Vert^2}{2\alpha}. $$
The exact characterization of $d_J^{ p^*}(u^* - \alpha \eta^*, u^*)$ depends on the properties of the functional and maybe  on $u^*$ itself. For $J$ being Frechet-differentiable with Lipschitz-continuous (or H\"older-continuous) derivative, it is always quadratic in $\alpha$, hence the estimate is of higher order in $\alpha$.  For the nonsmooth functionals like total variation or the $\ell^1$-norm the situation is different, at a first glance it cannot be expected that  $d_J^{ p^*}(u^* - \alpha \eta^*, u^*)$
is of higher order in $\alpha$. However, in such situations we can even have $d_J^{ p^*}(u^* - \alpha \eta^*, u^*) = 0$ for $\alpha$ small, e.g. in $\ell^1$ regularization if the support of $\eta^*$ is contained in the support of $u^*$.

The opposite question of weaker estimates arises if $u^*$ does not satisfy the source condition $p^* = F^*z^*$. In this case approximate source conditions are used, which measure the deviation from the source condition. A frequently used concept is the so-called {\em distance function} 
$$ D_\rho(p^*) = \inf \{ \Vert F^* z - p^* \Vert~|~ z \in {\cal V}, \Vert z \Vert \leq \rho \}, $$
which is useful in particular under strong convexity assumptions and allows to build a theory in a similar way by optimizing the value $\rho$ that finally appears in the error estimate. 
For functionals not being strictly convex and in particular the one-homogeneous cases like total variation a reformulation in terms of a dual problem is more suitable as seen in \cite{burger2016large}. There the measure
$$ e_{\alpha,\nu}(p^*) = \inf_{z \in {\cal V}} \nu J^*\left( \frac{F^* z - p^*}\nu \right)  + \frac{\alpha}{2} \Vert z \Vert^2 $$
was used to derive estimates. One observes some duality to the concept of distance functions, noticing that for $J$ being a norm in Banach space we just have
$$ e_{\alpha,\nu}(p^*) = \alpha \inf \{ \Vert z \Vert~|~ z \in {\cal V}, \Vert F^* z - p^* \Vert_*  \leq \nu \},$$
where $\Vert \cdot \Vert_*$ is the dual norm to $J$. 
 It was also shown that approximate source conditions are inherently related to the case of large noise, which is particularly relevant for stochastic models like white noise having non-finite variance (cf. \cite{bissantz2007convergence,burger2016large,kekkonen14}). 

While the literature was focused on asymptotic results for a long time, the specific shape of solutions at fixed positive $\alpha$ became a more attractive topic in the last two decades. In order to understand this issue a better understanding of the range condition for the regularization method is needed, which means the source condition $p^* = F^* z^*$ in the case of variational regularization. Since $F$ is modelled as a smoothing operator in inverse problems, $F^*$ is smoothing as well, which implies that the source condition is an abstract smoothness condition. However, the smoothness is rather indirect, since it concerns the subgradient $p^*$ and not directly $u^*$. Various results on the structure of minimizers, from sparsity properties for $J=\ell^1$ or its counterpart in the space of measures to total variation and staircasing phenomena can be found in literature (cf. \cite{caselles2007discontinuity,chambolle2010introduction}). 

Another issue that found strong recent interest is debiasing, since in the case of large noise the bias caused by the regularization term (and the large value of $\alpha$ that is needed to achieve stability)  spoil the possible quality of regularized solutions.  The influence of bias can also be seen from the term depending on $\Vert z^* \Vert$ in the error estimates, and in practice it is often observed that the reconstruction of the subgradient is better than the one of the primal solution due to bias. 
First debiasing methods (also called refitting) appeared in $\ell^1$ regularization, where in a first step the variational regularization is used and in a second step a simple least-squares problem is used on the support obtained from the first step, sometimes also with a sign constraint as obtained from the subgradient in the first step (cf. \cite{deledalle,lederer2013trust}). This approach can be translated to a more general two-step approach for 
debiasing as worked out in \cite{brinkmann2017bias}, which computes
$$ \Phi_\alpha(v) = \text{arg}\min \{ d_J^{p_\alpha}(u,u_\alpha) ~|~ u_\alpha \in \Phi^0_\alpha(v) \} ,$$
with $\Phi^0_\alpha$ being the regularization operator from the variational regularization method. 

Another approach effectively leading to debiasing, but also with other advantages, are iterative regularization methods such as the Bregman iteration (cf. \cite{osher2005iterative}). In the case of a quadratic functional, it can be formulated as an augmented Lagrangian method for computing the $J$-minimizing solution of $Fu=v$, i.e. 
\begin{align*}
u^{k+1} &\in \text{arg}\min_u  \frac{1}2 \Vert F u - v^k \Vert^2 + \alpha J(u) \\ 
v^{k+1} &= v^k + v - Fu^{k+1} , 
\end{align*}
with $v^0 =v$.  To have a suitable generalization also for other loss functionals this can be reformulated as 
\begin{align*}
u^{k+1} &\in \text{arg}\min_u  \frac{1}2 \Vert F u - v \Vert^2 + \alpha d_J^{p^k}(u,u^k) \\ 
p^{k+1} &=p^k + \frac{1}\alpha F^*(v - Fu^{k+1}) \in \partial J(u^{k+1}) . 
\end{align*}
The regularization parameter in this case is not $\alpha$, which is to be chosen rather larger in order to achieve good results, but the number of iterations carried out. Due to the variational structure in each iteration step, variational methods can be employed to prove well-definedness of the regularization operator, convergence, and error estimates. We refer to \cite{osher2005iterative,bur07,benningburger2018} for a detailed discussion of such iterative approaches and their analysis. Let us finally mention that in this respect there is another relation to machine learning, since Bregman iterations for $\ell^1$ regularizations have been developed further recently for the training of sparse deep neural networks and their architecture design (cf. \cite{bungert1,bungert2}). 

\section{Variational Regularization and Generalization}

In this final part we discuss some possible relations between the setup in machine learning and the above results on 
variational regularization theory. In particular we highlight some connections between the typical error measures used in the
two fields, namely generalization errors on the one hand and Bregman distances on the other.

\subsection{Error Decomposition and Generalization Error}
%
%$$ f(x;\theta) = \frac{1}N \sum_{i=1}^N \theta_i^2 \sigma(\theta_i^1 \cdot x + \theta_i^0). $$
%
%$$ F(x) =  \int \theta^2(a,b) \sigma(a \cdot x + b) ~d\mu(a,b)$$ 
%
%$$ F^* = \text{arg}\min_{F \in {\cal F}} ~ \expec_{(x,y) \sim P}(\ell(F(x),y) )$$ 
%
%$$ \expec_{ P}(\partial_F \ell(F^*(x),y)(F(x) - F^*(x) ) = 0$$ 
%
%
%$$ d(f,F^*) =  \expec_{ P}(\ell(f(x;\theta),y) - \ell(F^*(x),y) - \partial_F \ell(F^*(x),y)(f(x;\theta) - F^*(x) ) )$$
%
Let us return to the setup of machine learning with the minimization of the empirical risk with a convex loss $\ell$, taking the viewpoint that the ideal solution is the function $f^*$. While we have seen that 
naturally Bregman distances are estimated in the theory of variational regularization, the generalization error
$$ G =  {\mathbb E}_{(x,y)\sim {\mathbb P}}(\ell(f(x;\theta),y)) -  {\mathbb E}_{(x,y)\sim{\mathbb P}^N}(\ell(f(x;\theta),y)) $$
is the commonly used quantity in machine learning.

In order to understand the connections to Bregman distances consider an ideal solution $f^* \in {\cal F}$ minimizing the population risk, i.e.,
$$ f^* \in \text{arg}\min_{f \in {\cal F}} {\mathbb E}_{(x,y)\sim {\mathbb P}}(\ell(f(x),y)) = \text{arg}\min_{f \in {\cal F}} R^*(f). $$
Since the population risk is convex with respect to $f$, we conclude $0 \in \partial R^*(f)$, which implies
$$ d_{R^*}^0(f(.,\theta),f^*) = {\mathbb E}_{(x,y)\sim {\mathbb P}}(\ell(f(x;\theta),y)) - {\mathbb E}_{(x,y)\sim {\mathbb P}}(\ell(f^*(x),y)).$$ 
The latter can be decomposed in a similar spirit to the error decomposition in \cite{berner2021}
\begin{align*}
d_{R^*}^0(f(.,\theta),f^*) =& ~ {\mathbb E}_{(x,y)\sim {\mathbb P}}(\ell(f(x;\theta),y)) -  {\mathbb E}_{(x,y)\sim{\mathbb P}^N}(\ell(f(x;\theta),y))  + \\
         & ~ {\mathbb E}_{(x,y)\sim{\mathbb P}^N}(\ell(f(x;\theta),y)-\ell(f^*(x),y)) + \\
				& ~ {\mathbb E}_{(x,y)\sim{\mathbb P}^N}(\ell(f^*(x),y)) - {\mathbb E}_{(x,y)\sim{\mathbb P}}(\ell(f^*(x),y)).
\end{align*}
We see that the Bregman distance is decomposed into three parts: in addition to the generalization error in the first line, we have an approximation error in the second line (or rather a term that can be controlled with an approximation error in standard spaces) and a sampling error in the last line. The approximation error can be estimated beforehand or is often even negligible, since overparametrized models such as deep neural networks can usually be trained to have  ${\mathbb E}_{(x,y)\sim{\mathbb P}^N}(\ell(f(x;\theta),y) \approx 0$ and the second part is nonpositive. Moreover, the last term vanishes on expectation over the sampling if ${\mathbb P}^N$ is obtained from i.i.d. samples. Thus, in order to control the expected Bregman distance, the most important term is indeed the expected generalization error. 

\subsection{Estimates with Operator Errors and Generalization }

Errors due to sampling are effectively related to operator errors in inverse problems, which we see also from Example \ref{radonexample}, where effectively the operator $F$ is replaced by an operator $\tilde F$ being the concatenation of $F$ with a random sampling operator. 
%For the sake of a concise presentation we ignore data errors here, i.e. $v=v^* =Fu^*$.  
Moreover, we assume again a source condition of the form $p^* = F^* z^* \in \partial J(u^*)$.

The generalization error in this notation is given by (noticing that we might need to use different norms for the two terms)
$$ G(u) = \Vert F u - v \Vert^2 -  \Vert \tilde F u - \tilde v \Vert^2. $$
Hence, let us start again with the optimality condition of a regularized solution
$$ u_\alpha \in \text{arg}\min_u \frac{1}2 \Vert \tilde F u - \tilde v \Vert^2 + \alpha J(u), $$
which is given by
$$ \tilde F^* (\tilde F u_\alpha - \tilde v) + \alpha p_\alpha = 0, \qquad p_\alpha \in \partial J(u_\alpha). $$
Rewriting to 
$$  F^* F(  u_\alpha - u^*) + \alpha (p_\alpha - p^*)=  F^* ( F u_\alpha - v )  -  \tilde F^* (\tilde F u_\alpha - \tilde v) - \alpha F^* z^*, $$
we are in position to derive the kind of estimate we are after. A duality product with $u_\alpha-u^*$ and several applications of Young's inequality imply
$$ \frac{1}4 \Vert F(  u_\alpha - u^*) \Vert^2 + \alpha d_J^{p_\alpha,p^*}(u_\alpha,u^*) \leq \alpha^2 \Vert  z^* \Vert^2 + \Vert \tilde F u^* - \tilde v \Vert^2 + \frac{1}2 G(u_\alpha). $$
In the case of consistent data, such as obtained from sampling $F$, we further have $\tilde v = \tilde F u^*$, i.e., we obtain in particular
$$   d_J^{p_\alpha,p^*}(u_\alpha,u^*) \leq \alpha  \Vert  z^* \Vert^2 + \frac{1}{2 \alpha} G(u_\alpha) . $$
Thus, the error in the Bregman distance is controlled by the systematic error and the generalization error. 

\subsection{Regularized Risk Minimization Problems}

The above arguments can be extended to convex risk minimization problems of the form
$$ D_\alpha(\theta) = {\mathbb E}_{(x,y)\sim{\mathbb P}^N}(\ell(f(x;\theta),y)) + \alpha J(\theta)). $$
For simplicity we assume that the model $f$ is linear, i.e. $f(x;\theta) = (F\theta)(x)$ with a linear operator $F$ mapping to an appropriate function space ${\cal F}$, and $\ell$ is the squared Euclidean norm.  Consequently  we will consider $F$ as a bounded linear operator from some parameter space $\Theta$ to $L^2_{\mathbb P}(\Omega)^m$ for some domain $\Omega \subset \R^d$. The ideal solution $\theta^*$ is a minimizer of the population risk
$$ R(\theta) = {\mathbb E}_{(x,y)\sim{\mathbb P} } (\Vert (F\theta)(x) -y \Vert^2). $$

With this setup, the regularization operator is given by 
\begin{equation} \label{Phirisk} \Phi_\alpha({\mathbb P}^N) = \text{arg}\min_\theta {\mathbb E}_{(x,y)\sim{\mathbb P}^N}(\frac{1}2 \Vert (F\theta)(x) -y \Vert^2 + \alpha J(\theta)).  \end{equation}
Moreover, the source condition becomes 
\begin{equation} \label{sourcerisk} p^* = F^*z^* \in \partial J(\theta^*) \quad \text{with} \quad z^* \in L^2_{\mathbb P}(\Omega)^m
\end{equation} 
Similar to the reasoning in the previous section we can use the optimality condition
$$  {\mathbb E}_{(x,y)\sim{\mathbb P}^N}( \langle (F\theta_\alpha)(x) -y, F \theta' \rangle= + \alpha p_\alpha = 0, \quad p_\alpha \in \partial J(\theta_\alpha) $$
for all $\theta' \in \Theta$ to derive the following result:

\begin{thm}
Let $\theta_\alpha \in \Phi_\alpha({\mathbb P}^N) $ be defined by \eqref{Phirisk} and let the source conditon \eqref{sourcerisk} be satisfied. Then for appropriate $p_\alpha \in \partial J(u_\alpha)$ the estimate
\begin{align*}
&\frac{1}4 {\mathbb E}_{(x,y)\sim{\mathbb P}}( \Vert (F\theta_\alpha)(x) - (F\theta^*)(x) \Vert^2)+ \alpha d_J^{p_\alpha,p^*}(\theta_\alpha,\theta^*) \leq \\ & \qquad \qquad\qquad\frac{1}2 G(\theta_\alpha) + \alpha^2 \Vert z^* \Vert^2 + \mathbb E_{(x,y)\sim{\mathbb P^N}}(\Vert (F\theta^*)(x) -y \Vert^2).
\end{align*} 
with the generalization error
$$ G(\theta_\alpha)  = {\mathbb E}_{(x,y)\sim{\mathbb P}}(\frac{1}2 \Vert (F\theta_\alpha)(x) -y \Vert^2) - {\mathbb E}_{(x,y)\sim{\mathbb P}^N}(\frac{1}2 \Vert (F\theta_\alpha)(x) -y \Vert^2) . $$
\end{thm}

%
%\subsection{Theorems etc.}
%The statements of theorems, lemmas etc. are set in italics, but you can use $\backslash$emph\{\} to emphasize text therein. In definitions, only the term being defined is italicized. Remarks and examples are set in roman type.
%
%\begin{defin}\label{def-label}
%A system $S$ is said to be \emph{self-extensional} if $S \in B$.
%\end{defin}
%
%\begin{thm}[Maximum Principle; see also {\cite[Theorem 5]{Shchepin}}]
%If (\ldots), then the following conditions are equivalent:
%\begin{enumerate}
%\item first item,
%\item second item.
%\end{enumerate}
%\end{thm}
%
%\begin{proof} Observe that
%\begin{align}\label{E:1}
%AAAAAAAAAA &= BBBBBBBBBBB\notag\\
%&\quad + CCCCCCCCCC\notag\\
%&= DDDDDDDDDDDDD.
%\end{align}
%Now apply induction on $n$ to \eqref{E:1} and use Definition \ref{def-label}.
%\end{proof}
%
%If you want to put the end-of-proof sign after a (non-numbered) formula, use $\backslash$qedhere:
%
%\begin{proof} This follows from
%\begin{gather*}
%BBBBBBBBBBBBBBBB=CCCCCCCCCCCCC,\\
%DDDDDDD-EEEEEEEEEE=0.\qedhere
%\end{gather*}
%\end{proof}
%
%\begin{prob}
%Is AAAA true?
%\end{prob}
%
%\begin{rem}
%Remarks are unnumbered.
%\end{rem}
%
%\begin{mainthm}
%Here comes the statement of a numbered theorem with a fancy name.
%\end{mainthm}
%%
%%\subsection{Figures}
%Figures must be prepared and included as EPS or PDF files. All figures will be printed black and white; the colors will only appear in the online version.

\begin{acknowledgments}
This work was partly supperted by ERC via Grant EU FP7 ERC Consolidator
Grant 615216 LifeInverse, by the German Ministry of Science and Technology (BMBF) under
grant 05M2020 - DELETO, and by the EU under grant 2020 NoMADS - DLV-777826.
\end{acknowledgments}

\end{document}